\documentclass[sigconf, nonacm]{acmart}

\usepackage{float}
\usepackage{latexrelease}  % 加载LaTeX版本兼容包
\usepackage{amsmath}       % 常见的数学宏包，可能涉及计数器定义
\usepackage{appendix}
\usepackage{tabularx}
\usepackage{subcaption}
\usepackage{multirow} 
\newcommand\vldbdoi{XX.XX/XXX.XX}

\newcommand\vldbavailabilityurl{https://github.com/Jamoremore/MoLER}
\author{Hao Lin$^{1}$}
\affiliation{%
  \institution{Southeast University}
  \city{Nanjing}
  \country{China}
  \postcode{211189}
}
\email{220230828@seu.edu.cn}

\author{Peitong Xie$^{1}$}
\affiliation{%
  \institution{Nanyang Technological University}
  \city{Singapore}
  \country{Singapore}
}
\email{xiep0004@e.ntu.edu.sg}

\author{Jingxue Chen}
\affiliation{%
  \institution{Wired Product Operation Division, ZTE Corporation}
  \city{Nanjing}
  \country{China}
}
\email{chen.jingxue@zte.com.cn}

\author{Jie Lin}
\affiliation{%
  \institution{Wired Product Operation Division, ZTE Corporation}
  \city{Nanjing}
  \country{China}
}
\email{lin.jie1@zte.com.cn}

\author{Qingkun Tang$^\dagger$}
\affiliation{%
  \institution{Wired Product Operation Division, ZTE Corporation}
  \city{Nanjing}
  \country{China}
}
\email{tang.qingkun@zte.com.cn}
\thanks{$^{\mathrm{1}}$These authors contributed equally to this work.\\$^\dagger$Corresponding author. }

\author{Qianchun Lu}
\affiliation{%
  \institution{Wired Product Operation Division, ZTE Corporation}
  \city{Nanjing}
  \country{China}
}
\email{}

\begin{document}
\title{Domain-Aware RAG: MoL-Enhanced RL for Efficient Training and Scalable Retrieval}

\begin{abstract}

Retrieval-Augmented Generation (RAG) systems rely heavily on the retrieval stage, particularly the coarse-ranking process. Existing coarse-ranking optimization approaches often struggle to balance domain-specific knowledge learning with query enhencement, resulting in suboptimal retrieval performance. To address this challenge, we propose MoLER, a domain-aware RAG method that uses MoL-Enhanced Reinforcement Learning to optimize retrieval. MoLER has a two-stage pipeline: a continual pre-training (CPT) phase using a Mixture of Losses (MoL) to balance domain-specific knowledge with general language capabilities, and a reinforcement learning (RL) phase leveraging Group Relative Policy Optimization (GRPO) to optimize query and passage generation for maximizing document recall. A key innovation is our Multi-query Single-passage Late Fusion (MSLF) strategy, which reduces computational overhead during RL training while maintaining scalable inference via Multi-query Multi-passage Late Fusion (MMLF). Extensive experiments on benchmark datasets show that MoLER achieves state-of-the-art performance, significantly outperforming baseline methods. MoLER bridges the knowledge gap in RAG systems, enabling robust and scalable retrieval in specialized domains.
\end{abstract}

\maketitle

% %%% do not modify the following VLDB block %%
% %%% VLDB block start %%%
% \pagestyle{\vldbpagestyle}
% \begingroup\small\noindent\raggedright\textbf{PVLDB Reference Format:}\\
% \vldbauthors. \vldbtitle. PVLDB, \vldbvolume(\vldbissue): \vldbpages, \vldbyear.\\
% \href{https://doi.org/\vldbdoi}{doi:\vldbdoi}
% \endgroup
% \begingroup
% \renewcommand\thefootnote{}\footnote{\noindent
% This work is licensed under the Creative Commons BY-NC-ND 4.0 International License. Visit \url{https://creativecommons.org/licenses/by-nc-nd/4.0/} to view a copy of this license. For any use beyond those covered by this license, obtain permission by emailing \href{mailto:info@vldb.org}{info@vldb.org}. Copyright is held by the owner/author(s). Publication rights licensed to the VLDB Endowment. \\
% \raggedright Proceedings of the VLDB Endowment, Vol. \vldbvolume, No. \vldbissue\ %
% ISSN 2150-8097. \\
% \href{https://doi.org/\vldbdoi}{doi:\vldbdoi} \\
% }\addtocounter{footnote}{-1}\endgroup

% %%% VLDB block end %%%

\ifdefempty{\vldbavailabilityurl}{}{
\vspace{.3cm}
\begingroup\small\noindent\raggedright\textbf{Artifact Availability:}\\
The source code, data, and/or other artifacts have been made available at \url{\vldbavailabilityurl}.
\endgroup
}

\begin{figure*}
    \centering
    \includegraphics[width=\linewidth]{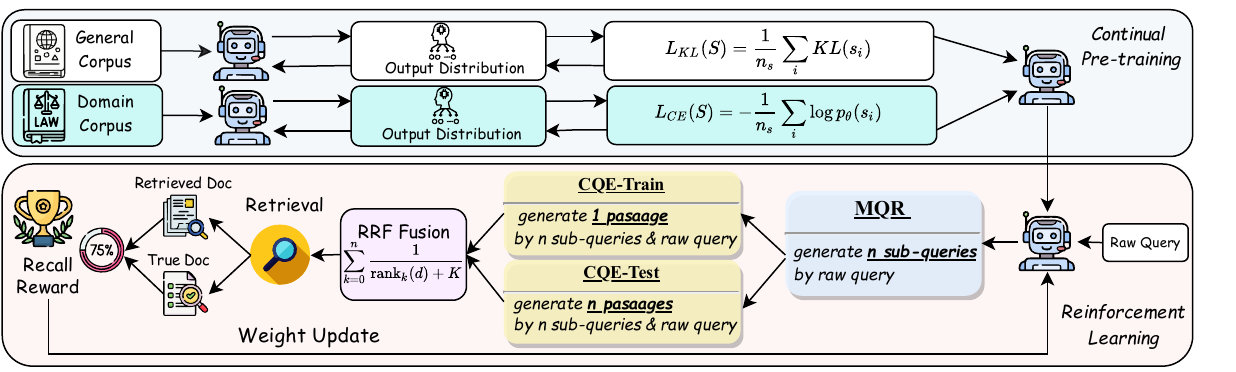}
    \caption{The MoLER framework's training and inference pipeline. During training, the model first undergoes CPT using the MoL approach, which balances domain-specific (CE loss) and general knowledge (KL divergence) corpora. In the RL phase, the MQR generates diverse queries, which are consolidated into a single synthetic passage via CQE to optimize recall performance. During inference, MQR generates multiple queries, each of which is processed independently to produce a passage. The retrieval results for these passages are then fused using RRF to enhance final recall performance.}
    \label{fig:framework}
\end{figure*}

\section{Introduction}
Retrieval-Augmented Generation (RAG) has emerged as a pivotal framework in natural language processing (NLP), comprising three core stages: retrieval, conditioning on retrieved documents and input, and generation\cite{Patrick2020}. While RAG significantly enhances the generation capabilities of large language model (LLM), its efficacy is heavily contingent upon the quality of the retrieval phase. During retrieval, coarse-ranking methods initially estimate query-document similarity to generate an initial recall order, followed by reranking to refine the results~\cite{nogueira2020}. Given the critical dependency of reranking on coarse-ranking outcomes~\cite{Zeng2022}, substantial research has focused on improving coarse-ranking methodologies. This work specifically targets advancements in coarse-ranking techniques to address the limitations of existing approaches.

To enhance coarse-ranking performance, various query augmentation methods have been proposed as critical strategies~\cite{chen2025,xia2025}. These methods, including query expansion~\cite{Bhogal2007}, Query2Doc~\cite{wang2023query2doc}, and multi-query generation~\cite{Jia2024}, aim to improve the relevance and diversity of retrieved documents by reformulating or expanding the original query. For instance, query expansion replaces keywords with synonyms to generate alternative queries~\cite{Dipasree2013,Xu2017}, while Query2Doc leverages LLM to pre-generate pseudo-passages for retrieval. However, these approaches often overlook the LLM's contextual understanding of documents, a critical gap in RAG scenarios where models typically operate in domains with limited prior knowledge~. For example, synonym-based expansion shows diminishing returns in dense semantic retrieval~\cite{karpukhin2020dense}, and Query2Doc's pseudo-passage generation lacks explicit alignment with the model's comprehension of document content. These limitations underscore the need for a methodology that directly optimizes retrieval performance from the perspective of document relevance, rather than relying on heuristic or template-based augmentation. Jiang et al.~\cite{jiang2025} proposed the DeepRetrieval method, which employs reinforcement learning (RL) to train an LLM capable of generating and rewriting augmented queries with good retrieval performance. However, the LLM was not explicitly trained to utilize stronger query augmentation techniques such as instruction expansion or pre-answering.

% Xiao et al.~\cite{bge_embedding} proposed C-Pack, advancing coarse ranking performance through training general Chinese text embedding models through a three-stage training pipeline (pre-training, contrastive learning, and task-specific fine-tuning). Shao et al.~\cite{shao2025reasonir} proposed an automated pipeline called REASONIR-SYNTHESIZER to generate training data that requires reasoning, and used this data to train retriever which significantly improves retrieval performance on RAG. Limitation of training embedding model and retriever is that the method is highly dependent on the quality and diversity of the training data.

% To address these challenges, we propose MoLER, a domain-aware RAG method that leverages MoL-Enhanced RL to optimize retrieval efficiency and scalability. MoLER integrates multi-query expansion and LLM pre-answering techniques, leveraging RL to optimize retrieval performance in an end-to-end manner. This approach is grounded on two critical premises: (1) continuous learning to deepen the model's understanding of retrieval documents~\cite{mol2025}, and (2) utilizing RL to activate the LLM's learned knowledge for effective query augmentation~\cite{jiang2025,nguyen2025}. By systematically addressing these requirements, MoLER bridges the gap between conventional augmentation techniques and the evolving demands of RAG systems, particularly in scenarios where retrieval effectiveness is directly proportional to the model's domain expertise.

To address these challenges, we propose MoLER, a domain-aware RAG framework that leverages MoL-Enhanced RL to optimize retrieval efficiency and scalability. MoLER integrates multi-query expansion and LLM pre-answering techniques, grounded in three core principles: (1) deepening the model’s comprehension of retrieved documents via continual learning~\cite{mol2025}; (2) achieving more efficient query enhancement through query expansion and question pre-answering; and (3) further activating the LLM’s inherent knowledge reservoir through RL to enable effective query augmentation. MoLER systematically bridges the gap between conventional enhancement techniques and the evolving demands of RAG systems, demonstrating significant advantages in scenarios where retrieval performance highly depends on the model’s domain-specific expertise.

For the first requirement, we employ the MoL (Mixture of Losses) training methodology, which distinguishes between domain-specific and general corpora~\cite{mol2025}. Domain corpora are trained using cross-entropy (CE) loss, while general corpora utilize KL divergence. Empirical results demonstrate that this approach effectively enhances domain expertise while preserving general capabilities.

\begin{figure*}[htbp]
    \centering
    \begin{subfigure}[b]{\linewidth}
        \includegraphics[width=\linewidth]{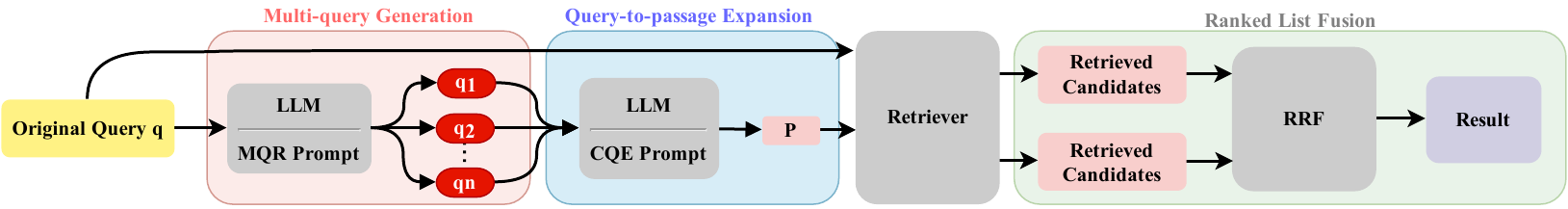}
        \caption{Training: Multi-query Single-passage Late Fusion Retrieval}
        \label{fig:sub1_1}
    \end{subfigure}
    \hfill
    \begin{subfigure}[b]{\linewidth}
        \includegraphics[width=\linewidth]{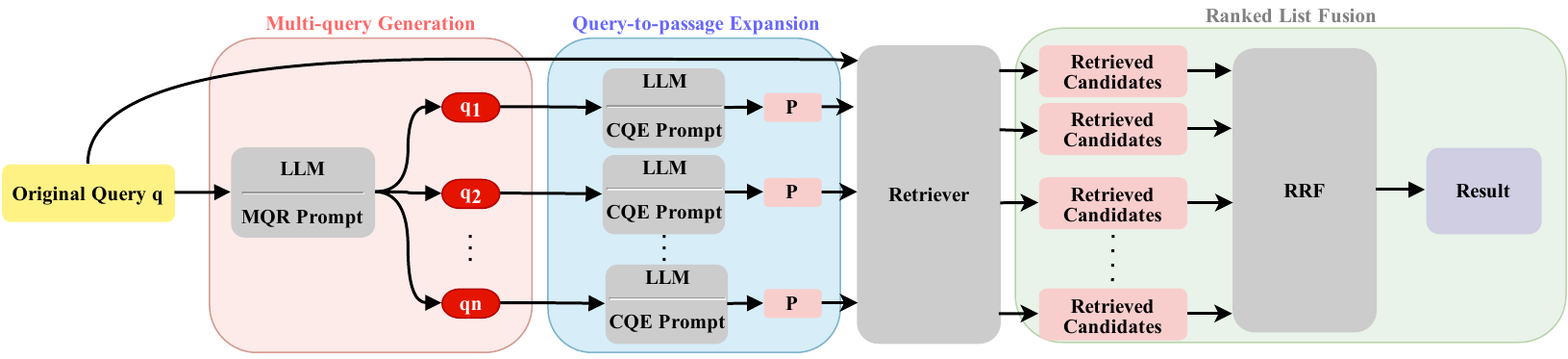}
        \caption{Testing: Multi-query Multi-passage Late Fusion Retrieval}
        \label{fig:sub2_1}
    \end{subfigure}
    \caption{The figure illustrates the distinction between MSLF and MMLF in MoLER's retrieval pipeline. (a) Training Phase (MSLF): Multiple queries generated via MQR are consolidated into a single synthetic passage using CQE. (b) Inference Phase (MMLF): Each query is processed independently to produce a distinct passage, and retrieval results are fused using RRF. This approach maximizes recall during inference by leveraging diverse query-document interactions. The design balances training efficiency (MSLF) and inference effectiveness (MMLF), ensuring optimal retrieval performance without compromising computational resources.}
    \label{fig:totalRAG_pipeline}
\end{figure*}

For the second and third requirements, we apply the Group Relative Policy Optimization (GRPO)~\cite{guo2025deepseek} based RL algorithm to optimize end-to-end recall. Although state-of-the-art methods such as multi-query multi-passage late fusion (MMLF)\cite{kuo2025mmlf} are assumed to exhibit strong scalability in retrieval performance with the number of generated queries and passages, this property is not formally validated in the original MMLF work and remains an open research question. The MMLF paper itself acknowledges this as a limitation, noting that computational efficiency during training and inference becomes prohibitive when generating multiple passages. To address this critical trade-off, we propose Multi-query Single-passage Late Fusion (MSLF), which reduces training complexity by consolidating multiple queries into a single synthetic passage during RL optimization. Crucially, our experiments demonstrate that the scalability benefits of MMLF. Specifically, a log-linear relationship between retrieval performance and the number of generated queries, are preserved in inference even after MSLF training. This strategy enables us to verify the scalability hypothesis of MMLF while maintaining training efficiency, ensuring the model retains MMLF's query expansion capabilities in deployment without incurring excessive computational costs during RL training.

We conduct rigorous experiments on NFCORPUS and SCIFACT, two representative retrieval datasets, to evaluate MoLER's performance. The results show that MoLER consistently outperforms baseline methods across all datasets, achieving statistically significant improvements in document recall. The superior recall performance underscores the importance of end-to-end optimization from the perspective of document relevance, particularly in scenarios where LLMs lack domain-specific knowledge.

Our contributions are as follows:
\begin{itemize} 
\item We identify the critical limitation of existing query augmentation methods: their failure to account for the LLM's contextual understanding of documents, which is particularly vital in RAG scenarios where models often lack domain expertise.
\end{itemize} 
\begin{itemize} 
\item A novel method (MoLER) that bridges the gap between traditional query augmentation techniques and dynamic RAG requirements through RL-based end-to-end optimization.
\end{itemize} 
\begin{itemize} 
\item The MSLF method, which enables efficient RL training via single-passage generation while maintaining MMLF-based scalability during inference, achieving a critical balance between training efficiency and inference effectiveness in downstream domains.
\end{itemize} 
\begin{itemize} 
\item We demonstrate through extensive experiments on benchmark datasets (e.g., NFCORPUS, SCIFACT) that MoLER achieves significant improvements in retrieval performance, validating its effectiveness in bridging the knowledge gap in RAG systems. 
\end{itemize} 

\section{Related Work}
\paragraph{Query Agumentation} Query augmentation aims to reformulate and agunent user's input query which may be ambigous and cover insufficient dimensions. Query expansion is an earlier proposed method that replaces keywords in the original query with synonyms~\cite{Bhogal2007}. Although it demonstrated good performance in early sparse retrievers based on keyword matching~\cite{Dipasree2013,Xu2017}, query expansion has shown limited performance improvements with the rise of dense retrievers based on semantic
matching~\cite{karpukhin2020dense,Xiong2020}. Query2Doc leverages LLMs to firstly generate a pseudo-passage based on oringinal query and then uses this enriched context to retrieve~\cite{wang2023query2doc}. Mill proposed a Query-Query-Document method which utilizess LLMs to decomposes the original query into multi-dimensional sub-queries, pre-answer each sub-queries and concatenate  pseudo-passages into a continuous text for retrieval~\cite{Jia2024}. MMLF employs Multi Query Retriever (MQR) to generate diverse sub-queries and Combined Query Expansion (CQE) to expand them into contextual passages utilizes~\cite{kuo2025mmlf}, and fusing rankings via reciprocal rank fusion (RRF)  instead of concatenate pseudo-passages~\cite{Cormack2009}.

\paragraph{Continuous Pre-training} Traditional CPT~\cite{Huang2019,cossu2022} methods typically mix general and domain-specific corpora to train models to mitigate catastrophic forgetting. However, determining the optimal mixing ratio is challenging and requires extensive ratio experiments. Song et al. propose a dual-objective optimization strategy during domain-specific tuning to address catastrophic forgetting, using regularized loss for general data and cross-entropy loss for domain data~\cite{Shezheng2025}. However, they did not provide the optimal ratio for general and domain data. Chen et al.~\cite{mol2025} introduce a novel dual-loss architecture, MOL, to address these issues: general corpora are trained with KL divergence loss to preserve foundational capabilities, while domain-specific corpora are trained with CE loss to enhance specialized knowledge. They explicitly propose that a nearly 1:1 ratio of domain to general corpora effectively balances training outcomes, avoiding the waste of computational resources.

\paragraph{RL for LLM Query Agumentation} Recent works leverage RL to optimize query generation in retrieval tasks. Jiang et al.\cite{jiang2025} proposed DeepRetrieval, an RL framework that trains LLMs to generate/rewrite queries using task-specific reward metrics. Xiao et al.\cite{bge_embedding} introduced C-Pack, a three-stage pipeline (pre-training, contrastive learning, fine-tuning) to enhance Chinese text embeddings for coarse ranking. Shao et al.~\cite{shao2025reasonir} developed REASONIR-SYNTHESIZER, an automated data generation pipeline for training retrievers with reasoning-required queries. These approaches demonstrate the potential of RL and data synthesis in improving retrieval performance but focus on different aspects of the problem.

\begin{table}
    \centering
    \caption{MQR Prompt}
    \label{tab:mqr_prompt}
    \begin{tabularx}{\columnwidth}{|X|}
        \hline
        \textbf{MQR prompt} \\ \hline
You are an AI language model assistant. Your task is to generate exactly \{cnt\} different versions of the given user question to retrieve relevant documents from a vector database. By generating multiple perspectives on the user question, your goal is to help the user overcome some of the limitations of the distance-based similarity search.\\\\
Original question: {query}\\\\
Format your response in plain text as:\\\\
\{example\}
\\ \hline
    \end{tabularx}
\end{table}

\begin{table}
    \centering
    \caption{CQE Prompt}
    \label{tab:cqe_prompt}
    \begin{tabularx}{\columnwidth}{|X|}
        \hline
        \textbf{CQE prompt} \\ \hline
Please write a passage to answer the following user questions simultaneously.\\\\
Question 1: \{original\_query\}\\\\
Question 2: \{sub\_query\_1\}\\\\
...\\\\
Question n: \{sub\_query\_n+1\}\\\\
Format your response in plain text as:\\\\
Passage:
\\ \hline
    \end{tabularx}
\end{table}

\section{Method}

For improving the performance of coarse ranking, this paper proposes a novel training framework named MoLER. As shown in \autoref{fig:framework}, MoLER consists of two core stages: a CPT stage based on MoL and a retrieval optimization stage based on RL.

\begin{enumerate}
\item \textbf{CPT with MoL.} We adopt a dual-loss architecture in which the training corpus is divided into two categories: domain documents and general documents. For domain knowledge, we apply CE loss to enhance domain-specific knowledge learning based on retrievable documents. For general knowledge, we use the KL divergence loss, treating the LLMs reasoning corpus as general knowledge to preserve the model’s general language capabilities.

\item \textbf{RL based fine-tuning.} In this stage, RL is employed to distill compressed knowledge into specialized document retrieval capability. First, an MQR Prompt is used to semantically expand the raw query; then, a CQE Prompt is used to generate a preliminary answer. Finally, the preliminary answer and the original query are combined using RRF to further optimize the retrieval performance on private-domain documents.
\end{enumerate}

This section first introduces the document retrieval strategy adopted by MoLER, followed by a detailed explanation of the algorithmic design and process implementation of the CPT stage and the RL stage.

\subsection{Redesigned RAG Pipeline}
\label{rag_pipeline}

% The redesigned RAG pipeline of MoLER is the "instruction expansion–pre-answer guidance–reciprocal rank fusion" process. Instruction expansion enables the model to understand a query from multiple perspectives, while pre-answer guidance improves the alignment of retrieved results with the target answer. 
MoLER enhances retrieval performance through the same three-step process as~\cite{kuo2025mmlf}: instruction expansion, pre-answer guidance, and reciprocal rank fusion.

\begin{enumerate}

\item \textbf{Instruction Expansion:} Using a zero-shot prompting approach, the original query $q$ is expanded into $n$ semantically related sub-queries $\{q_1, q_2, \dots, q_n\}$ with varied perspectives and richer details. These form the expanded query set $\mathcal{Q}$, providing multi-view retrieval entry points.

\item \textbf{Pre-Answer Guidance:}  Traditional RAG methods directly retrieve documents using the original query, which can lead to retrieval bias when the query is ambiguous. MoLER introduces a pre-answer mechanism. During the training phase, all sub-queries are merged to generate a unified pseudo-document $p$. During the testing phase, a pseudo-document $p_i$ is generated for each $q_i$ to serve as contextual enhancement for retrieval. Specifically, to enhance both instruction expansion and pre-answer generation capabilities with similar token consumption while effectively reducing the number of dialogues required for training, we adopt an MSLF strategy during training and an MMLF strategy during testing (as shown in \autoref{fig:totalRAG_pipeline}).
% Traditional RAG methods directly retrieve documents using the original query, which can lead to retrieval bias when the query is ambiguous. MoLER introduces a pre-answer mechanism: in the training phase, all sub-queries are merged to generate a unified pseudo-document $p$; in the testing phase, a pseudo-document $p_i$ is generated for each $q_i$, serving as contextual enhancement for retrieval. Namely, to balance training efficiency and model performance, we adopt the MSLF
% strategy during training, and the MMLF strategy during testing, as illustrated in \autoref{fig:totalRAG_pipeline}.

\item \textbf{Reciprocal Rank Fusion:} The retrieval results from the original query and the sub-queries are fused using RRF, as follows:
\begin{itemize}
    \item Compute similarity scores between $q$ and the coarse-ranked documents to obtain the similarity ranking list $L_0$;
    \item Compute similarity scores between each pseudo-document $p_i$ (corresponding to $q_i$) and the coarse-ranked documents to obtain the ranking lists $L_i$;
    \item Compute the final RRF score and ranking for the original query $q$ using:
    \begin{equation}
    s(q) = \sum_{k=0}^n \frac{1}{\text{rank}_k(d) + K}
    \end{equation}
    where $\text{rank}_k(d)$ denotes the rank of document $d$ in $L_k$, and $K=60$ is the standard RRF constant~\cite{kuo2025mmlf}.
\end{itemize}

\end{enumerate}

\begin{table*}
    \centering
    \caption{This table compares the retrieval performance of various methods on the NFCORPUS and SCIFACT datasets. The proposed MoLER method (Qwen3-1.7B+MoL+GRPO) achieves state-of-the-art results, significantly outperforming baseline methods such as Raw Query, Q2D, CoT, and LC-MQR.} 
    \label{tab:main_results_overall}
    \begin{tabular}{l *{2}{c c}} 
    \toprule
        \multirow{2}{*}{Method} & 
        \multicolumn{2}{c}{NFCORPUS} & 
        \multicolumn{2}{c}{SCIFACT}  \\
        \cmidrule(lr){2-3} \cmidrule(lr){4-5} 
         & Recall@1k & nDCG@10 & Recall@10 & nDCG@10\\
        \midrule
        Raw Query & 52.95 & 21.69& 59.82 & 46.57  \\
        Qwen3-32B+Q2D & 60.06 & \underline{25.15}& 78.29 & 62.20  \\
        Qwen3-32B+CoT& 59.30 & 24.17 & 71.98 & 57.89   \\
        Qwen3-32B+LC-MQR& 58.10 & 24.48 &73.36 &57.97    \\
        Qwen3-32B+MMLF& \underline{60.87} & 24.97 & \underline{79.26} & \underline{62.55}    \\
        Qwen3-0.6B+MoL+Dr.GRPO & 59.72 & 23.18 & 72.96 & 55.43 \\
        Qwen3-0.6B+MoL+GRPO & 60.52 & 23.59 & 77.73 & 61.00\\
        Qwen3-1.7B+MoL+Dr.GRPO  & 60.19 & 24.38 & 77.47 & 60.90  \\
        Qwen3-1.7B+MoL+GRPO  & \textbf{61.42} & \textbf{25.44} & \textbf{79.69} & \textbf{62.59} \\
    \bottomrule
    \end{tabular}
\end{table*}

\subsubsection{Key Prompt Strategies}
To implement the redesigned RAG pipeline, MoLER uses two critical prompt strategies, MQR and CQE, that enable efficient query augmentation and contextual alignment. Below, we detail their design and implementation.

The central idea of the MQR prompt is to generate multiple reformulated queries that mitigate semantic drift issues inherent in vector similarity search. Unlike the design in ~\cite{kuo2025mmlf}, we introduce two custom parameters: $cnt$, denoting the number of expanded queries, and $example$, which provides task-specific examples depending on the query number. Table~\ref{tab:mqr_prompt} shows the general format of the MQR prompt.

Following \cite{kuo2025mmlf}, we also adopt the CQE prompt. Unlike simple sub-query expansion, the CQE prompt generates a passage that simultaneously answers both the original query and the sub-queries. The corresponding template is provided in Table~\ref{tab:cqe_prompt}. The key innovation lies in its parameterized structure where the choice of 
$n$ (number of sub-queries) determines the underlying learning framework: when 
$n>1$, the CQE prompt operates under the MSLF strategy during training, generating a pseudo-document that simultaneously addresses multiple reformulated queries to enhance contextual alignment. Conversely, during inference, the prompt switches to the MMLF strategy by iterating over each expanded query (i.e., $n=1$ per iteration) to generate a distinct pseudo-document for each sub-query. This approach ensures that the system leverages diverse query-document interactions during inference while maintaining computational efficiency in training.

\subsection{Continual Pre-training}
When applying RL to LLMs, training from scratch proves highly inefficient, whereas CPT mitigates this by reducing RL exploration costs through domain-specific knowledge assimilation. Traditional CPT often encounters the dilemma of "domain overfitting" and "general capability dilution" when injecting domain-specific knowledge into LLMs, leading to a degradation in their general-purpose abilities and impairing comprehension of user queries. Therefore, it is crucial to develop CPT methods that enable LLMs to acquire domain knowledge while preserving their general capabilities. To address this, MoLER adopts the MoL dual-loss architecture~\cite{mol2025}, enabling collaborative optimization that strengthens domain knowledge while maintaining general capabilities through differentiated loss design.

Specifically, the training corpus is divided into a domain-specific corpus $C_d$ and a general-domain corpus $C_g$, optimized jointly with CE loss and KL divergence loss, respectively.

For a sequence $s$ from the domain-specific corpus $C_d$, the CE loss is defined as:
\begin{equation}
\mathcal{L}_{CE}(s) = -\frac{1}{n_s} \sum_i \log p_\theta(s_i)
\end{equation}
where $n_s$ is the sequence length, and $p_\theta(s_i)$ denotes the probability of generating token $s_i$ under parameters $\theta$.

For a sequence $s$ from the general-domain corpus $C_g$, the KL loss is defined as:
\begin{equation}
\mathcal{L}_{KL}(s) = \frac{1}{n_s} \sum_i \text{KL}[p_\theta \| p_0](s_i)
\end{equation}

Following~\cite{mol2025}, we employ an optimal 1:1 corpus ratio in LoRA fine-tuning to avoid degradation of generalization ability due to excessive domain data, thereby ensuring balanced performance.

\begin{figure*}
    \centering
    % Subfigure 1: Add width parameter, adjust label position
    \begin{subfigure}[b]{0.49\linewidth}
        \includegraphics[width=\linewidth]{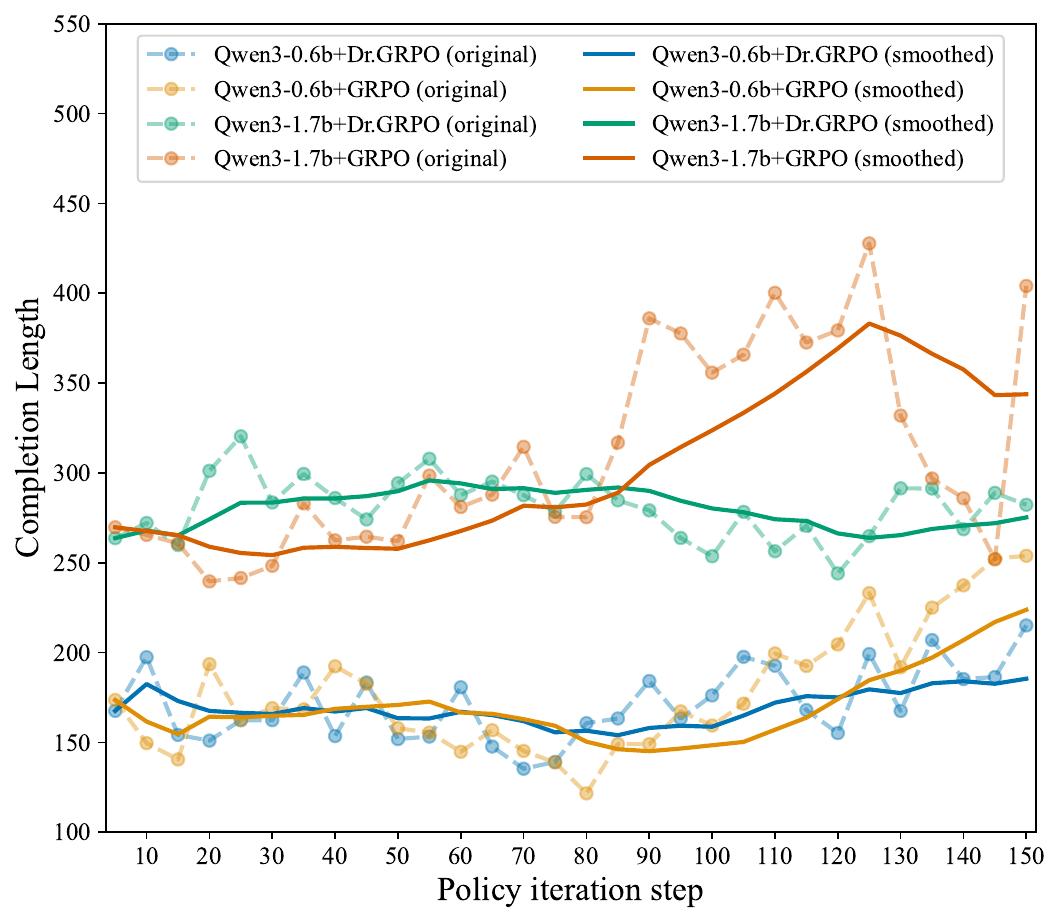}
        \caption{NFCORPUS}
        \label{fig:sub1}
    \end{subfigure}
    \hfill
    \begin{subfigure}[b]{0.49\linewidth}
        \includegraphics[width=\linewidth]{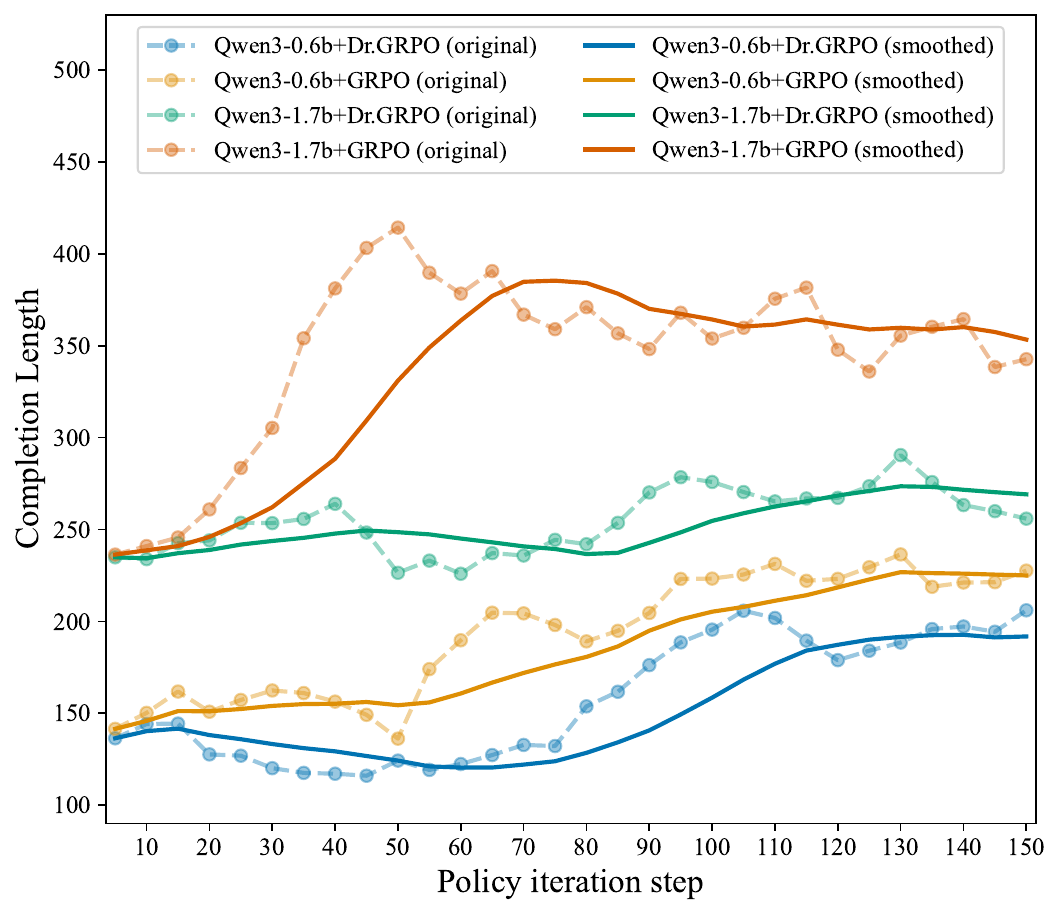}
        \caption{SCIFACT}
        \label{fig:sub2}
    \end{subfigure}
    
    \caption{Comparison of response lengths between GRPO and Dr.GRPO after multi-step training using Qwen3-0.6B and Qwen3-1.7B models on (a) NFCORPUS and (b) SCIFACT datasets. The results show that Dr.GRPO consistently generates shorter completion lengths across both model scales and datasets.}
    \label{fig:completion}
    
\end{figure*}

\subsection{Reinforcement Learning }

The core challenge in RAG tasks is to generate retrieval queries that align closely with user intent, thereby maximizing the recall rate for private-domain documents. Conventional supervised fine-tuning (SFT) relies heavily on manually annotated "query–document" pairs, which are scarce and costly to produce in practice. To overcome this, MoLER employs an unsupervised RL approach using the GRPO-based algorithm to directly optimize the probability distribution of retrieval strategies, with retrieval recall serving as the reward signal.

To assess the generalization capability of the proposed framework across different RL paradigms, we conduct a comparative analysis between GRPO and its representative variant, Dr.GRPO.

\subsubsection{GRPO and Dr.GRPO Algorithms}

GRPO is a policy optimization algorithm for multi-candidate output scenarios. It replaces the independent critic model in PPO (Proximal Policy Optimization)~\cite{schulman2017proximal} with group-wise relative advantage estimation, significantly reducing computational overhead. However, GRPO suffers from two types of bias:

\begin{enumerate}
\item \textbf{Response length bias:} The normalisation term $1/|o_i|$ reduces the gradient penalty for longer incorrect responses (negative advantage) while amplifying the positive gradient for shorter correct ones (positive advantage). This bias encourages increasingly verbose outputs, especially when responses are incorrect.
\item \textbf{Problem difficulty bias:} The advantage estimate $\hat{A}_{i} = \frac{R_i - \text{mean}(R)}{\text{std}(R)}$ assigns higher optimization weight to low-variance problems (i.e., extremely easy or extremely difficult tasks).
\end{enumerate}

The GRPO objective function is:
\begin{equation}
\begin{aligned}
&\mathcal{J}_{\text{GRPO}}(\pi_\theta) = \mathbb{E}_{q,\{o\}} \frac{1}{G} \sum_{i=1}^G \sum_{c} \frac{1}{|o_{i,c}|} \sum_{t=1}^{|o_{i,c}|}   \\
&\quad\quad\quad\quad\quad\quad \Bigg( \frac{\pi_\theta}{\pi_{\theta_{\text{old}}}} \hat{A}_{i}-\beta D_{\text{KL}}[\pi_\theta \| \pi_{\text{ref}}]\Bigg).
\end{aligned}
\end{equation}
The objective function is formulated to optimize the policy $\pi_{\theta}$ by maximizing the expected cumulative advantage across multiple query generation and pre-answering. Here, $c$ indexes the components of the retrieval pipeline, specifically the MQR and CQE prompts, which correspond to distinct stages in the query augmentation process. For each sample $i$ and component $c$, $o_{i,c}$ represents the model's output generated under prompt $c$, such as sub-queries (for MQR) or pseudo-passages (for CQE). 

To mitigate the response length and problem difficulty biases in GRPO and achieve unbiased optimization, Dr.GRPO~\cite{liu2025understanding} removes the $1/|o_i|$ term and the $\text{std}(R)$ normalization term, thereby decoupling gradient updates from response length while maintaining consistent optimization weights across problems. The revised objective function is:
\begin{equation}
\begin{aligned}
\mathcal{J}_{\text{Dr.GRPO}} \sim\,&\mathbb{E}_{q,\{o\}} \frac{1}{G} \sum_{i=1}^G \sum_{c}\sum_{t=1}^{|o_{i,c}|} \\&\Bigg( \frac{\pi_\theta}{\pi_{\theta_{\text{old}}}} \hat{A}_{i}-\beta D_{\text{KL}}[\pi_\theta \| \pi_{\text{ref}}]\Bigg), \\
\end{aligned}
\end{equation}

\noindent where $\quad \tilde{A}_{i} = R_i - \text{mean}(R)$.

The framework employs the MSLF strategy during RL training to reduce computational costs. While MMLF is utilized during inference to maximize retrieval effectiveness, the training phase benefits from MSLF's streamlined approach of generating a single pseudo-passage from multiple sub-queries. This design reduces the number of model interactions from 
$n+1$ (where $n$ denotes the number of query expansions) to 2 during policy rollouts, significantly accelerating training without compromising final performance. Empirical results demonstrate that MSLF-based RL training can still effectively enhance MMLF's retrieval capabilities during inference.

\begin{table*}
\centering
\caption{Comparison of base models and MoLER-trained models under nonthinking and thinking Modes. The results demonstrate that MoLER (Qwen3-0.6B/1.7B+MoL+Dr.GRPO) achieves comparable performance in both reasoning modes, with marginal gains in thinking mode across all metrics.}
\label{tab:think-results}
\begin{tabular}{l *{2}{c c}} 
\toprule
\multirow{2}{*}{Method} & 
\multicolumn{2}{c}{NFCORPUS} & 
\multicolumn{2}{c}{SCIFACT}  \\
\cmidrule(lr){2-3} \cmidrule(lr){4-5} 
 & Recall@1k & nDCG@10 & Recall@10 & nDCG@10\\
\midrule
Qwen3-0.6B+nonthinking   & 57.73 & 22.02 & 63.21 & 49.80 \\
Qwen3-0.6B+thinking   & 58.44 & 22.60 & 67.68 & 51.82 \\
Qwen3-0.6B+MoL+Dr.GRPO+nonthinking & \underline{59.55} & \underline{23.18} & \underline{72.96} & \underline{55.43} \\
Qwen3-0.6B+MoL+Dr.GRPO+thinking   & \textbf{59.72} & \textbf{23.61} & \textbf{73.10} & \textbf{56.30} \\
\midrule
Qwen3-1.7B+nothinking & 59.29 & 23.36 & 72.26 & 57.76 \\
Qwen3-1.7B+thinking  & 59.49 & 24.25 & 73.19 & 58.49\\
Qwen3-1.7B+MoL+Dr.GRPO+nothinking & \underline{60.19} & \underline{24.38} & \underline{77.47} & \underline{60.90} \\
Qwen3-1.7B+MoL+Dr.GRPO+thinking  & \textbf{60.57} & \textbf{24.65} & \textbf{77.69} & \textbf{61.57} \\
\bottomrule
\end{tabular}
\end{table*}

\begin{figure*}
    \centering
    % Subfigure 1: Add width parameter, adjust label position
    \begin{subfigure}[b]{0.49\linewidth}
        \includegraphics[width=\linewidth]{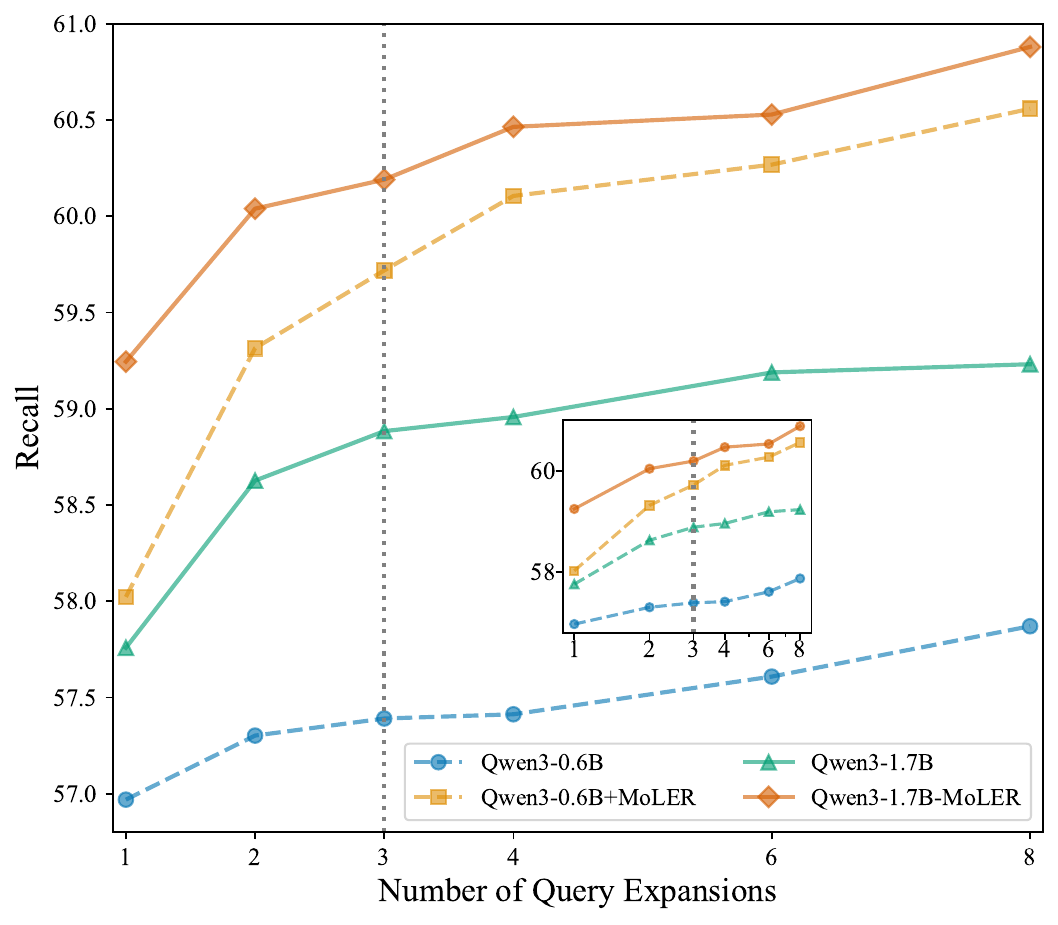}
        \caption{NFCORPUS}
        \label{fig:sub1}
    \end{subfigure}
    \hfill
    \begin{subfigure}[b]{0.49\linewidth}
        \includegraphics[width=\linewidth]{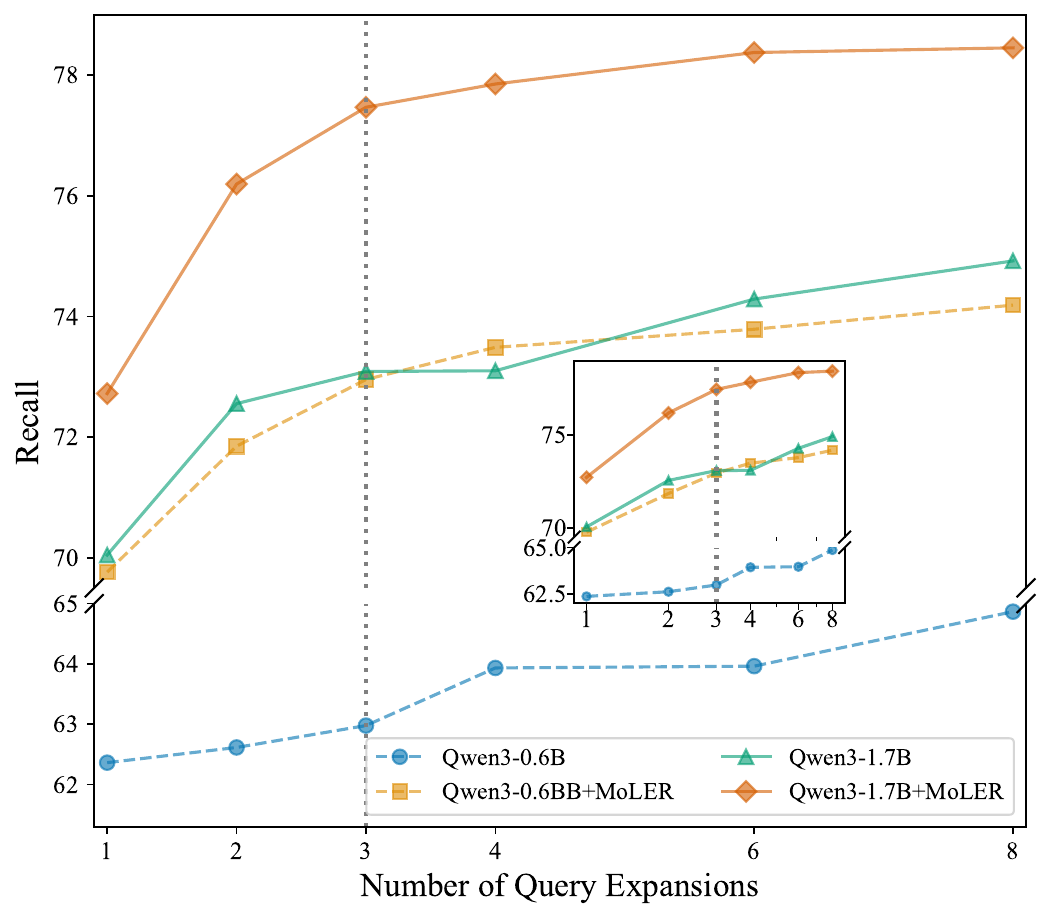}
        \caption{SCIFACT}
        \label{fig:sub2}
    \end{subfigure}
    
    \caption{Performance comparison between MoLER-enhanced models and base models across different query expansion counts on (a) NFCORPUS and (b) SCIFACT datasets. Results demonstrate that Qwen3-0.6B+MoLER achieves performance comparable to Qwen3-1.7B base model on SCIFACT, highlighting MoLER's effectiveness in enabling smaller models to achieve larger model performance. The logarithmic scaling subplots (shown as insets in the main figure) reveal near-linear relationships between recall improvements and query expansion counts, confirming adherence to scaling laws.}
    \label{fig:number_base}
    
\end{figure*}

\section{Experiments}\label{sec:experiments}
This section systematically evaluates the performance of the proposed MoLER framework on private retrieval tasks. We begin by establishing the experimental setup, including the selection of base models, datasets, and evaluation metrics that form the foundation of our empirical analysis. Our evaluation then proceeds with the main results, comparing MoLER against traditional RAG and various prompting strategies to demonstrate its overall performance advantages across different model scales. The analysis examines key characteristics including comparative analysis of GRPO variants, scalability of query expansion, and the effectiveness of nonthinking versus thinking modes. Subsequently, we conduct comprehensive ablation studies to dissect the contributions of individual components within the MoLER framework. These studies systematically investigate: (1) the impact of different retrieval fusion strategies (MSLF vs. MMLF); (2) the contribution of various retrieval augmentation methods and their combinations; and (3) the enhancement of retrieval capabilities brought by different continual pre-training approaches, specifically comparing our proposed MoL method against conventional domain-specific training strategies.

\subsection{Experimental Setup and Evaluation Metrics}
\subsubsection{Base Models and Embedding}
Considering the dual requirements of efficiency and performance in RAG scenarios, we selected the open-source Qwen3-0.6B and Qwen3-1.7B models as our base models to compare performance across different parameter scales. Furthermore, the embedding model used in all experiments is OpenAI's text-embedding-ada-002~\cite{openai_blog_embedding}.

\subsubsection{Datasets and Evaluation Metrics}

We selected two datasets from the BEIR benchmark~\cite{thakur2021beir} for methodological evaluation: NFCORPUS and SCIFACT. NFCORPUS focuses on biomedical question-answering retrieval, comprising 3,633 documents, 2,590 training queries, and 323 test queries, with an average of 38.2 relevant documents per query and an average query length of 3.3 words. SCIFACT is oriented toward scientific fact-checking, consisting of 5,183 paper abstracts, 809 training queries, and 300 test queries, with an average of only 1.1 relevant documents per query and an average query length of 12.37 words. Following the setup in~\cite{kuo2025mmlf}, NFCORPUS is evaluated using Recall@1k and nDCG@10, while Recall@10 and nDCG@10 are adopted for SCIFACT to achieve comparable recall levels. These two datasets cover distinct domains and query complexities, enabling an effective evaluation of the retrieval system’s robustness in a zero-shot setting.

\subsubsection{MoL Continual Pre-training}
During the continual learning phase, we employed the PEFT framework~\cite{mangrulkar2022peft} to conduct MoL training on the models using a domain-specific dataset and the general-purpose dataset Light-R1~\cite{lightr1proj}. To align with LLM pre-training principles, we transformed Light-R1 into unsupervised text using a dialogue template (e.g., concatenating turns with speaker roles as natural text sequences). This approach adheres to the token prediction objective of LLMs, ensuring the training process remains consistent with their pre-training paradigms. Moreover, the inclusion of dialogue data helps preserve the model's conversational understanding, which is critical for RAG systems involving dynamic user interactions. To improve parameter efficiency, we utilize Low-Rank Adaptation (LoRA) with a rank of 64~\cite{hu2022lora}. Throughout the MoL training, the model's context window is fixed at 8192 tokens to ensure effective coverage of document knowledge. Other hyperparameters are detailed in Appendix ~\ref{sec:appendix_mol}.

\begin{table*}
\centering
\caption{Comparison of MSLF and MMLF fusion strategies on NFCORPUS and SCIFACT datasets with fixed base models (Qwen3-0.6B/Qwen3-1.7B) and training strategies (MoL+Dr.GRPO/MoL). The results demonstrate that MMLF significantly outperforms MSLF in key metrics such as Recall@1k and Recall@10 across different model scales and training configurations, validating the effectiveness of MMLF in enhancing retrieval performance.}
\label{tab:fusion-results}
\begin{tabular}{l *{2}{c c}} 
\toprule
\multirow{2}{*}{Method} & 
\multicolumn{2}{c}{NFCORPUS} & 
\multicolumn{2}{c}{SCIFACT}  \\
\cmidrule(lr){2-3} \cmidrule(lr){4-5} 
 & Recall@1k & nDCG@10 & Recall@10 & nDCG@10\\
\midrule
Qwen3-0.6B+MoL+MSLF & 57.23 & 22.19 & 63.24 & 47.83 \\
Qwen3-0.6B+MoL+MMLF & 57.31 & 22.51 & 64.00 & 49.90 \\
Qwen3-0.6B+MoL+Dr.GRPO+MSLF &  \underline{58.02} & \textbf{23.62} &  \underline{70.77} &  \underline{53.82}  \\
Qwen3-0.6B+MoL+Dr.GRPO+MMLF  & \textbf{59.72} &  \underline{23.18} & \textbf{72.96} & \textbf{55.43} \\
\midrule
Qwen3-1.7B+MoL+MSLF & 58.37 & 23.82 & 70.34 & 54.19  \\
Qwen3-1.7B+MoL+MMLF &  \underline{59.41} & 24.14 & 73.22 & 57.29  \\
Qwen3-1.7B+MoL+Dr.GRPO+MSLF  & 59.27 & \textbf{24.59} & 75.16 &  \underline{57.45}  \\
Qwen3-1.7B+MoL+Dr.GRPO+MMLF   & \textbf{60.19} &  \underline{24.38} & \textbf{77.47} & \textbf{60.90}  \\
\bottomrule
\end{tabular}
\end{table*}

\begin{figure}
  \centering
  \includegraphics[width=\linewidth]{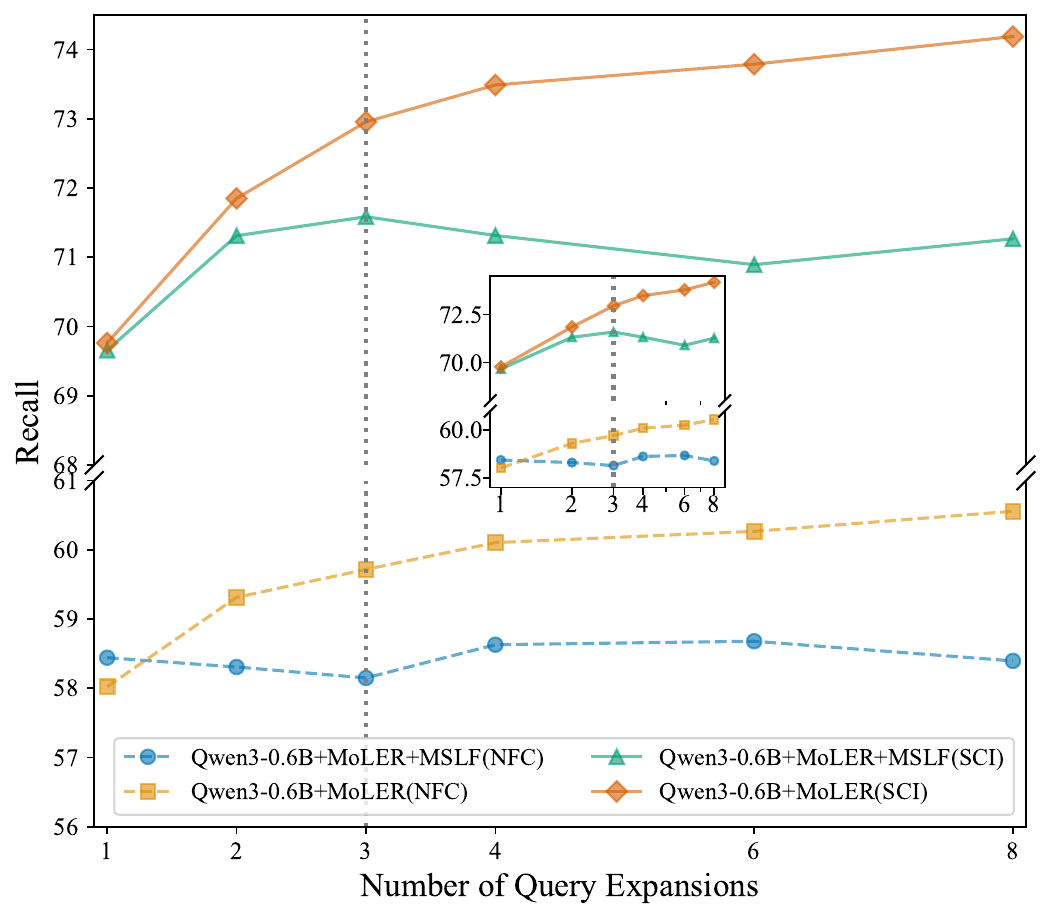}
   \caption{Performance comparison between MMLF and MSLF fusion strategies across varying query expansion counts on NFCORPUS and SCIFACT datasets using Qwen3-0.6B+MoLER models. The results demonstrate that MMLF exhibits logarithmic scaling in performance with increased expansion counts, while MSLF shows minimal improvement, indicating its limited scalability. Notably, training with MSLF (which uses only 3 fixed expansions) does not degrade MMLF's scaling capability during inference. This demonstrates the framework's effectiveness: the MSLF-based training strategy maintains computational efficiency during RL optimization while preserving MMLF's end-to-end scalability for optimal retrieval performance.}
  \label{fig:query_num_mslf}
\end{figure}

\subsubsection{GRPO-based post-training}
In the wake of MoL CPT, a GRPO-based post-training phase is introduced with the objective of further aligning the model's generative priors with the final objectives of the downstream retrieval task. The focal point of this stage is the construction of a reward signal that directly reflects retrieval performance. Specifically, the MSLF strategy (see Section~\ref{rag_pipeline} for details) is employed to guide the model in generating a series of query expansions for the original query, and further expanding these expanded queries into a pre-answer passage for the question. This passage, in conjunction with the original query, is then utilized to compute the cosine similarity between their embeddings and those of the original documents for the purpose of ranking. The results are consolidated into a final document list via RRF. The recall rate of relevant documents in this fused list is adopted as the core metric for the reward function. Further hyperparameters of the GRPO framework, incorporating the particular design and rollout Configuration of the reward function, are outlined in Appendix \ref{sec:appendix_grpo}. Furthermore, in the MSLF and MMLF strategies, the number of query expansions $n$ is explicitly set to 3 to balance computational efficiency and retrieval effectiveness.

\subsubsection{Evaluation Protocol and Implementation Details}
To ensure a fair comparison, enhancement techniques such as pseudo-relevance feedback, mutual verification~\cite{jia2024mill}, or few-shot prompting are not introduced during the evaluation. Unless otherwise specified, all models are run in nonthinking mode during both the RL training and testing phases, adopting the officially recommended decoding hyperparameters for Qwen3: a temperature of 0.7, top\_p of 0.8, and top\_k of 20~\cite{yang2025qwen3}. Additionally, MMLF is used for query augmentation Unless specified.

\begin{table*}
\centering
\caption{Ablation study of retrieval enhancement strategies on base models and MoLER framework. This table presents a comparative analysis of different retrieval augmentation methods (Raw Query, LC-MQR, Q2D) and their integration with the MoLER framework (MoL+Dr.GRPO+MSLF) on Qwen3-0.6B and Qwen3-1.7B models across the NFCORPUS and SCIFACT datasets. Key findings include: (1) The MoLER framework (MoL+Dr.GRPO) consistently improves retrieval metrics (Recall@1k, Recall@10, nDCG@10) over baseline methods across both model scales; (2) Combining MoL+Dr.GRPO with MMLF fusion achieves optimal performance, demonstrating that the synergistic design of "query expansion + question pre-answering" effectively compensates for individual strategy limitations. The results validate MoLER's superiority in enhancing retrieval robustness through end-to-end optimization of multi-perspective query-document interactions.}
\label{tab:ablation-enhance}
\begin{tabular}{l *{2}{c c}} 
\toprule
\multirow{2}{*}{Method} & 
\multicolumn{2}{c}{NFCORPUS} & 
\multicolumn{2}{c}{SCIFACT}  \\
\cmidrule(lr){2-3} \cmidrule(lr){4-5} 
 & Recall@1k & nDCG@10 & Recall@10 & nDCG@10\\
\midrule
Raw Query & 52.95 & 21.69& 59.82 & 46.57   \\
Qwen3-0.6B+LC-MQR  & 54.27 & 20.98 & 60.84 & 46.79\\
+MoL+Dr.GRPO  & 57.59 &  \underline{22.35} & 67.59 & 51.30 \\
Qwen3-0.6B+Q2D & 56.26 & 19.49 & 61.38 & 48.01 \\
+MoL+Dr.GRPO  &  \underline{58.11} & 20.41 &  \underline{70.44} &  \underline{54.14}  \\
Qwen3-0.6B+MMLF  & 57.73 & 22.02 & 63.21 & 49.80 \\
+MoL+Dr.GRPO  & \textbf{59.72} & \textbf{23.18} & \textbf{72.96} & \textbf{55.43} \\
\midrule
Qwen3-1.7B+LC-MQR  & 53.93 & 22.29 & 66.89 & 51.08 \\
+MoL+Dr.GRPO   & 55.16 & 22.51 & 68.97 & 53.16 \\
Qwen3-1.7B+Q2D  & 58.70 & 22.58 & 73.41 & 59.10  \\
+MoL+Dr.GRPO   &  \underline{59.31} & 23.32 &  \underline{75.20} &  \underline{59.21} \\
Qwen3-1.7B+MMLF  & 59.10 &  \underline{23.59} & 73.52 & 57.10  \\
+MoL+Dr.GRPO   & \textbf{60.19} & \textbf{24.38} & \textbf{77.47} & \textbf{60.90}   \\
\bottomrule
\end{tabular}
\end{table*}  

\subsection{Main Results}
\subsubsection{Overall Performance Comparison}
We validated the effectiveness of MoLER by comparing it against four categories of baseline methods: 1) \textbf{Raw Query}, which directly uses the initial query statement; 2) \textbf{Query2Doc (Q2D)~\cite{wang2023query2doc}}, which expands the query into a paragraph for retrieval; 3) \textbf{Chain of Thought (CoT)~\cite{cot_Jagerman2023}}, which transforms the query into an answer and its reasoning chain; and 4) \textbf{LangChain Multi-Query Retriever (LC-MQR)}, which generates multiple sub-queries and fuses the results using RRF.

Table~\ref{tab:main_results_overall} presents the overall experimental results. The data show that MoLER significantly enhances retrieval performance across Qwen3 models of different scales. Notably, Qwen3-1.7B+MoL+GRPO achieves the best performance on both tasks, with its recall and nDCG metrics ranking highest among all baselines. This combination even surpasses the Qwen3-32B+MMLF model, which has 18.82 times the number of parameters. Particularly on the recall metric, MoLER shows an average improvement of 0.49\% over its closest competitor, Qwen3-32B+MMLF, indicating that applying MoLER can produce substantial gains on smaller models.

\subsubsection{GRPO vs Dr.GRPO Algorithm Comparison}
As shown in Figure~\ref{fig:completion}, we compared the token consumption of different model sizes and RL algorithms during the training phase. After applying a moving average with a window size of 8, it is clearly observable that the token lengths during the Dr.GRPO training process are shorter and more stable. Given Dr.GRPO’s significant advantage in response length, and although its recall and nDCG performance may not consistently surpass that of GRPO, possibly due to the nature of our tasks, and we ultimately select Dr.GRPO as the primary reference algorithm for subsequent experiments.

% \begin{table}
%     \centering
%     \caption{Token Consumption across models} 
%     \label{tab:model_token_results}
%     \begin{tabular}{lccc}
%     \toprule
%         Model       & NFCORPUS& SCIFACT \\
%         \midrule
%         Qwen3-0.6B   & 1360.34 & 1247.96 \\
%         +MoL+GRPO & 2112.31 & 1376.31 \\
%         +MoL+Dr.GRPO  & 2008.48 & 1181.97 \\
%         \midrule
%         Qwen3-1.7B & 1965.15 & 1822.93  \\
%         +MoL+GRPO & 2681.48 & 1964.01 \\
%         +MoL+Dr.GRPO & 2229.02 & 1737.24 \\
%     \bottomrule
%     \end{tabular}
%     \caption{}
% \end{table}

\begin{table*}
\centering
\caption{Comparison of pre-training methods with varying training epochs on retrieval performance. This table evaluates the impact of excluding general-domain corpora versus domain-aware MoL training with 2/4 epochs on retrieval effectiveness under Dr.GRPO. Key findings include: (1) Increasing MoL training from 2 to 4 epochs consistently improves Recall@1k and nDCG@10; (2) The optimal performance is achieved with MoL+Dr.GRPO after 4 training epochs, demonstrating that MoLER enhances the model's ability to learn domain-specific patterns while maintaining robustness through the dual-loss architecture.}

\label{tab:pretrain-results}
\begin{tabular}{l *{2}{c c}} 
\toprule
\multirow{2}{*}{Method} & 
\multicolumn{2}{c}{NFCORPUS} & 
\multicolumn{2}{c}{SCIFACT}  \\
\cmidrule(lr){2-3} \cmidrule(lr){4-5} 
 & Recall@1k & nDCG@10 & Recall@10 & nDCG@10\\
\midrule
Qwen3-0.6B   & 57.73 & 22.02 & 63.21 & 49.80 \\
+Dr.GRPO  & 59.22 & \textbf{23.63} & 71.79 & 55.12 \\
+CE(4 epoch) & 57.02 & 22.44 & 64.00 & 49.90  \\
+CE(4 epoch)+Dr.GRPO  & \underline{59.32} & 22.84 & \underline{72.17} & \underline{55.19}  \\
+MoL(4 epoch) & 57.31 & 22.51 & 64.83 & 50.76 \\
+MoL(2 epoch)+Dr.GRPO  & 59.28 & \underline{23.18} & 72.12 & 54.99 \\
+MoL(4 epoch)+Dr.GRPO  & \textbf{59.72} & \underline{23.18} & \textbf{72.96} & \textbf{55.43} \\
\bottomrule
\end{tabular}
\end{table*}

\begin{figure}
  \centering
  \includegraphics[width=\linewidth]{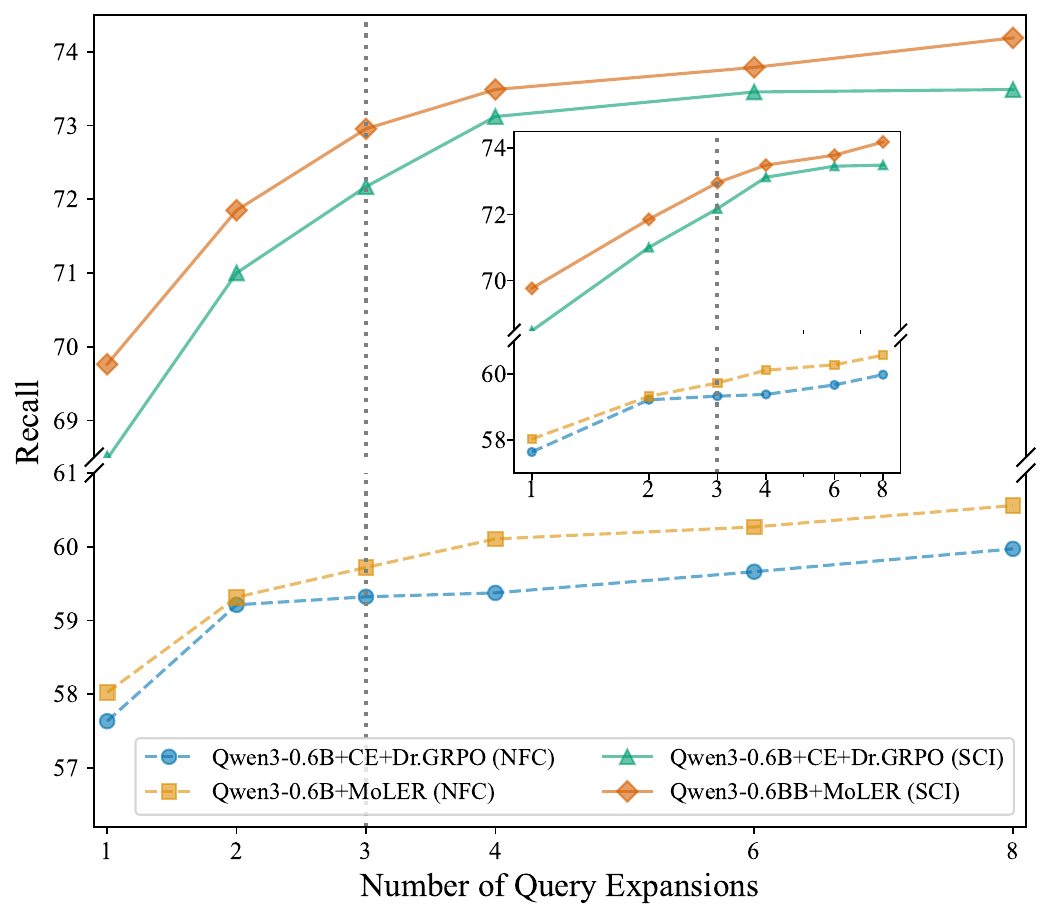}
  \caption{Impact of query expansion count on MoLER versus CE+Dr.GRPO performance across NFCORPUS and SCIFACT datasets. The figure compares Recall performance as a function of the number of query expansions for Qwen3-0.6B models trained with MoLER and CE+Dr.GRPO methods. Results demonstrate that MoLER consistently outperforms CE+Dr.GRPO across varying expansion counts, with both methods showing logarithmic scaling behavior.}
  \label{fig:query_num_ce}
\end{figure}

\subsubsection{Scalability Analysis of Query Expansion}

Figure~\ref{fig:number_base} presents the performance on MoLER-enhanced models and base models across different query expansion counts in MMLF. The results demonstrate two key findings. First, when the x-axis is scaled logarithmically, both~\autoref{fig:number_base} (a) and~\autoref{fig:number_base} (b) exhibit approximately linear growth patterns in recall performance. This linear relationship under log-scale suggests that recall improvement follows scaling laws with respect to the number of query expansions, where performance gains adhere to a power-law relationship. The scaling behavior is consistent across both NFCORPUS and SCIFACT datasets, demonstrating MoLER's scalable characteristics in different retrieval scenarios.

Second, on the SCIFACT dataset shown in~\autoref{fig:number_base} (b), Qwen3-0.6B+MoLER achieves recall performance comparable to the Qwen3-1.7B base model. This indicates that MoLER enables smaller parameter models to achieve performance levels similar to larger base models.

\subsubsection{Nonthinking vs Thinking Mode Analysis}
The experimental results in Table~\ref{tab:think-results} demonstrate that both base models and MoLER-trained models exhibit comparable performance in nonthinking and thinking modes, with marginal improvements observed in the latter. For instance, on the MoLER-trained Qwen3-0.6B model, the Recall@1k metric increases by only 0.17\% (from 59.55 to 59.72) and nDCG@10 by 0.43\% (from 23.18 to 23.61) when switching from nothink to thinking mode. These findings suggest two key insights: (1) CoT reasoning does not substantially enhance retrieval performance for knowledge gaps in the LLM, as the marginal improvements fall well below the gains achieved by MoLER's retrieval optimization strategies. (2) Adopting nonthinking mode for training and inference is computationally efficient and practical, as the minimal performance trade-off is outweighed by reduced latency and resource consumption.

\subsection{Ablation Study}

\subsubsection{Impact of Different Retrieval Fusion Strategies}

To examine the effectiveness of the proposed MMLF and MSLF strategies, we conduct comparative experiments under Dr.GRPO optimization. The results are presented in Table~\ref{tab:fusion-results}.

Experimental results confirm that employing the MSLF strategy during the RL phase is an effective approach to enhancing the performance of MMLF. Specifically, after applying the Dr.GRPO algorithm, which uses the MSLF-derived recall as its optimization objective, both the Qwen3-0.6B and Qwen3-1.7B models demonstrate significant improvements in recall and nDCG metrics for both MSLF and MMLF across the two datasets. This finding not only validates the effectiveness of this RL strategy but also proves that MoLER can effectively boost the model's performance in complex retrieval tasks. Furthermore, in line with expectations, MMLF's performance metrics consistently surpass those of MSLF, both before and after the RL phase. The reasoning is twofold: MSLF is selected during the RL phase primarily to improve training efficiency, whereas MMLF, by leveraging its "pre-answer fusion" strategy, is able to explore associated documents from more diverse perspectives, thereby achieving superior final performance.

Figure~\ref{fig:query_num_mslf} provides additional insights into the scaling behavior of these fusion strategies across different query expansion counts. The results reveal distinct scaling characteristics between MMLF and MSLF approaches. MMLF demonstrates clear scaling law adherence, with performance improving logarithmically as the number of query expansions increases on both NFCORPUS and SCIFACT datasets. In contrast, MSLF shows performance fluctuations without a clear scaling trend across different expansion counts. This behavioral difference can be attributed to MSLF's architectural constraint of utilizing only single-passage semantic similarity computation, which inherently limits its sensitivity to variations in query expansion count. The observed scaling patterns reinforce our design choice of using MSLF during the RL training phase while deploying MMLF for final inference, as this combination optimizes both training efficiency and inference performance.

\begin{figure}
  \centering
  \includegraphics[width=\linewidth]{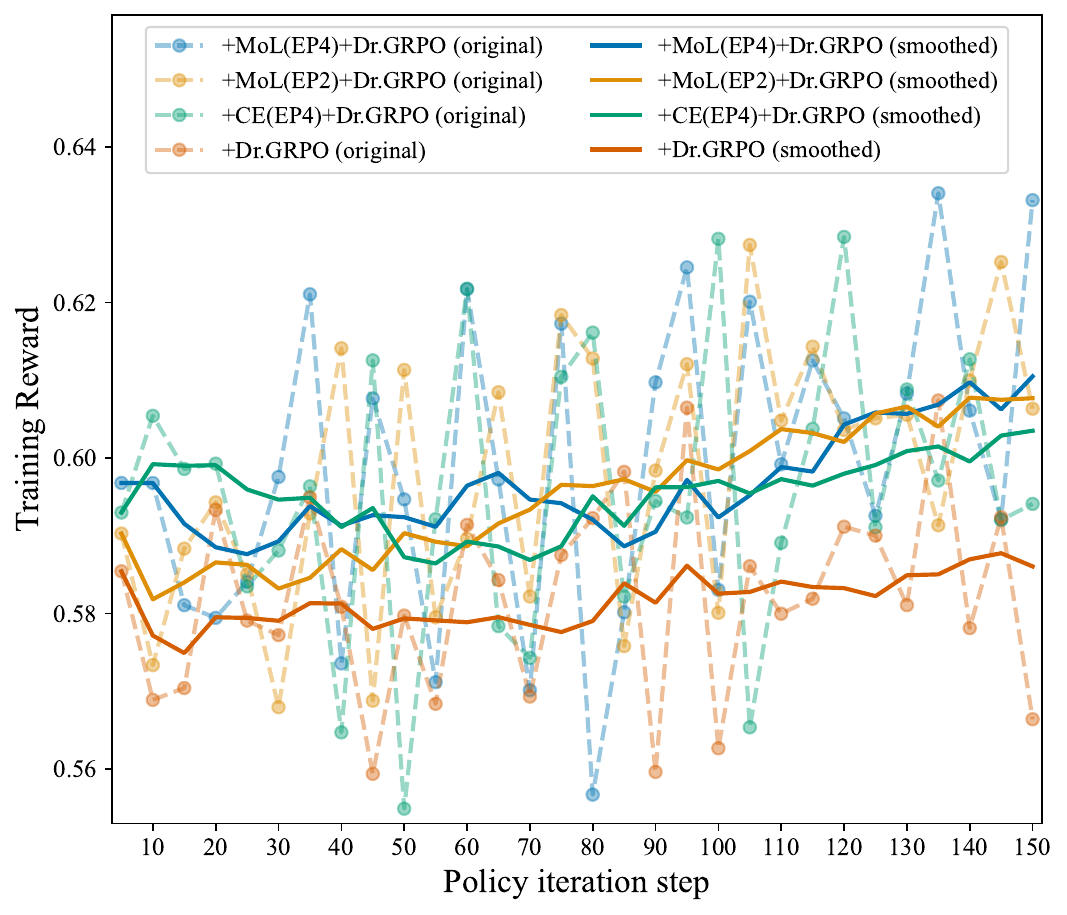}
  \caption{Convergence trends of Dr.GRPO reward curves on Qwen3-0.6B with varying pre-training strategies on NFCORPUS. The plot compares the RL (Dr.GRPO) reward progression during training, focusing on the impact of removing general corpora from MoL and adjusting the number of MoL training epochs (2 vs. 4). A moving average with a window size of 8 is applied for clarity. The results demonstrate that MoL pre-training with 4 epochs achieves the most stable and superior reward trajectory, highlighting its effectiveness in balancing domain knowledge acquisition and generalization.}
  \label{fig:pretrain_drgrpo}
\end{figure}

\subsubsection{Comparison of Retrieval Augmentation Methods}

To further analyze the contributions of different retrieval augmentation strategies, we compared the effects of Raw Query, LC-MQR, Q2D, and their combinations with MoL+Dr.GRPO. The results are shown in Table~\ref{tab:ablation-enhance}. The main findings are as follows.

\begin{enumerate}
\item For the Qwen3-0.6B model on the NFCORPUS dataset, using either LC-MQR or Q2D alone resulted in a decrease in nDCG, suggesting that relying on a single augmentation strategy may introduce bias. In contrast, this phenomenon is not observed with the Qwen3-1.7B model. We speculate this may be due to the smaller parameter count of the Qwen3-0.6B model, leading to weaker generation capabilities for query expansion and question pre-answering.
\item Regardless of the retrieval augmentation method used, the introduction of MoL+Dr.GRPO consistently led to significant improvements in retrieval performance, manifesting as higher Recall and nDCG, and demonstrating a consistent advantage over Raw Query.
\item When further combined with MMLF, performance reached its optimum. This validates that the "query expansion + question pre-answering" design, when used jointly, can compensate for the shortcomings of individual strategies, thereby achieving the best retrieval effectiveness.
\end{enumerate}

\subsubsection{Comparison of Different CPT Methods}

We further compared the impact of different CPT methods on the model's retrieval capability. The experimental results, shown in Table~\ref{tab:pretrain-results}, include comparisons of CE, MoL, and MoL with different training epochs. In particular, the CE method utilizes only domain-specific documents to be retrieved.

The experimental results in Table~\ref{tab:pretrain-results} demonstrate that MoLER exhibits robustness to variations in pre-training strategies, consistently achieving stable improvements across different CPT methods and the proposed MoL framework outperforms CE in both recall and nDCG metrics. This advantage is particularly pronounced when training is constrained to the same 4 epochs, underscoring the efficiency of MoL's dual-loss design in optimizing retrieval performance without compromising model versatility.

On the other hand, all models showed significant performance improvements after the introduction of Dr.GRPO, validating the effectiveness of the RL phase in improving retrieval capabilities. For MoL, training for 4 epochs performed better than for 2 epochs, which further corroborates the rationality of our chosen pre-training setup.

Additionally, Figure~\ref{fig:query_num_ce} provides further insights into the scaling behavior of MoLER compared to CE+Dr.GRPO across different numbers of query expansions. The results demonstrate that MoLER maintains consistent performance advantages over CE+Dr.GRPO across all expansion counts on both datasets. Notably, both methods exhibit similar logarithmic scaling patterns, confirming that the performance improvements follow theoretical scaling laws regardless of the pre-training strategy employed. This finding reinforces that MoLER's superior performance stems not only from its domain-aware pre-training approach but also from its inherent scalability characteristics, making it a robust choice for retrieval enhancement across different operational scales.

Figure~\ref{fig:pretrain_drgrpo} displays the convergence curves of different pre-training methods on the NFCORPUS dataset. For ease of observation, we compare the results after applying a moving average with a window size of 8. It can be observed that applying Dr.GRPO directly to the base model yields the most limited improvement. This is because, without the support of domain-specific knowledge, the model needs to spend more time exploring effective strategies. In contrast, after introducing MoL, whether for 2 or 4 epochs, the model exhibits a superior final performance. This indicates that MoL not only preserves general-purpose capabilities but also provides a better foundation for RL.

\section{DISCUSSION AND LIMITATIONS}

% The experimental results of this study unequivocally confirm that the proposed MoLER framework significantly enhances the performance of RAG systems. The two-stage approach, combining CPT with MoL and the joint application of RL with GRPO, have both proven to be highly effective. The core insight lies in the nature of the document-based knowledge repository: generating multiple passages during inference can introduce diverse retrieval perspectives, thereby effectively improving retrieval efficiency. This indicates that MSLF training optimizes computational efficiency while maintaining the scalability of MMLF; due to the inherent diversity of underlying documents ensuring performance robustness, its efficacy is maintained even with fewer expansions during the training phase. This design effectively balances the trade-off between training efficiency and inference effectiveness in domain-specific retrieval scenarios.

% Despite these promising results, one critical limitation remains: the inference latency associated with the MMLF strategy. During inference, the MMLF approach requires multiple forward passes through the language model to generate and process expanded queries and pre-answer passages, which increases computational overhead compared to single-query methods. Future work would focus on optimizing the inference pipeline through parallelization, lightweight query expansion modules, or compression techniques to reduce this latency while maintaining retrieval effectiveness.

Our experimental results demonstrate that the proposed MoLER framework significantly enhances retrieval performance in RAG systems. The two-stage approach integrating CPT with MoL and RL with GRPO proves highly effective.

Despite the significant achievements of MoLER, several critical limitations and avenues for future research warrant discussion. Firstly, acquiring and curating high-quality, human-annotated domain-specific data for the RL phase can be costly and time-consuming, particularly in highly specialized or proprietary domains. While MoLER effectively leverages such data, reducing reliance on extensive manual annotation or exploring semisupervised/unsupervised methods for domain knowledge acquisition will be important research directions.

Secondly, this study focuses mainly on text-based RAG systems. Extending MoLER to handle multimodal information will open new avenues for more comprehensive knowledge retrieval. Furthermore, exploring the integration of MoLER with structured knowledge bases or knowledge graphs could enhance its reasoning capabilities and provide more precise and verifiable answers beyond pure document retrieval. The principles of domain-aware learning and RL-based optimization developed in MoLER can be adapted to these complex data environments, thereby further solidifying its contribution to large-scale information management and retrieval within database systems.

\section{CONCLUSION}

In this paper, we address a critical limitation in existing RAG systems: the disconnect between query augmentation strategies and the direct optimization of retrieval performance. To bridge this gap, we introduced MoLER, a novel two-stage framework that enhances informative retrieval for domain-specific tasks. MoLER first employs MoL for CPT, enabling the model to acquire specialized knowledge without suffering from catastrophic forgetting. Subsequently, it utilizes GRPO-based RL to fine-tune the model's query and passage generation policy, directly maximizing document recall.

Our extensive experiments on the NFCORPUS and SCIFACT datasets demonstrate the effectiveness of the MoLER framework. The results show that even a compact 1.7B parameter model equipped with MoLER can significantly outperform strong baselines, including a much larger 32B parameter model using conventional augmentation techniques. This underscores MoLER's parameter efficiency and its ability to unlock the full potential of smaller models for complex retrieval tasks. Key contributions of our work are threefold: (1) deepen the model’s comprehension of retrieved documents through continual learning; (2) achieve more efficient query enhancement via multi-query expansion and LLM pre-answering; (3) further activate the LLM’s inherent knowledge reservoir through RL to enable effective query augmentation. Ultimately, MoLER provides a robust and scalable solution for building highly effective and efficient domain-adaptive RAG systems, paving the way for more powerful and knowledgeable AI applications.

\appendix
\section{Appendix} \label{sec:appendix}

\subsection{MoL Continual Training Hyperparameters}
\label{sec:appendix_mol}
Table~\ref{tab:mol_setup} presents the hyperparameter configuration adopted during the continual training of MoL. The choice of these hyperparameters aims to ensure training stability while enhancing the model’s generalization ability in cross-task transfer.

\subsection{GRPO-based Post-Training Hyperparameters}
\label{sec:appendix_grpo}
In the post-training phase, we applied the GRPO-based framework to further refine the retrieval and generation performance of the model. Table~\ref{tab:grpo_setup} summarizes the major hyperparameter settings, emphasizing robustness under adversarial scenarios. Meanwhile, Table~\ref{tab:grpo_rollout} provides the rollout configuration during inference, covering parameters such as parallelism strategy, number of generations, and temperature.

\begin{table}
  \centering
  \caption{Training Setup for MoL}
  \label{tab:mol_setup}
  \begin{tabular}{lc}
    \toprule
   Hyperparameter & Value \\
    \midrule
    global batch size & 128\\
    learning rate & 2e-4 \\
    LoRA rank & 64\\
    weight decay& 0.1 \\
  \bottomrule
\end{tabular}
\end{table}

\begin{table}
  \centering
  \caption{Training Setup for GRPO}
  \label{tab:grpo_setup}
  \begin{tabular}{lc}
    \toprule
   Hyperparameter & Value \\
    \midrule
    global batch size & 64\\
    gradient 
    learning rate & 1e-4 \\
    LoRA rank & 64\\
    weight decay& 0.1 \\
  \bottomrule
\end{tabular}
\end{table}

\begin{table}
  \centering
  \caption{Rollout Configuration for GRPO}
  \label{tab:grpo_rollout}
  \begin{tabular}{lc}
    \toprule
   Parameter & Value \\
    \midrule
    rollout backend & vLLM\\
    tensor parallel size & 1 \\
    data parallel size & 1 \\
    num generation & 8 \\
    max completion length & 8192 \\
    temperature & 0.9\\
  \bottomrule
\end{tabular}
\end{table}

\bibliographystyle{ACM-Reference-Format}
\bibliography{sample}

\end{document}